# Nonparametric Variational Inference

**Samuel J. Gershman**                                                                    SJGERSHM@PRINCETON.EDU

Department of Psychology, Princeton University, Green Hall, Princeton, NJ 08540 USA

**Matthew D. Hoffman**                                                                   MDHOFFMA@CS.PRINCETON.EDU

Department of Statistics, Columbia University, New York, NY 10027 USA

**David M. Blei**                                                                            BLEI@CS.PRINCETON.EDU

Department of Computer Science, Princeton University, 35 Olden St., Princeton, NJ 08540 USA

## Abstract

Variational methods are widely used for approximate posterior inference. However, their use is typically limited to families of distributions that enjoy particular conjugacy properties. To circumvent this limitation, we propose a family of variational approximations inspired by nonparametric kernel density estimation. The locations of these kernels and their bandwidth are treated as variational parameters and optimized to improve an approximate lower bound on the marginal likelihood of the data. Unlike most other variational approximations, using multiple kernels allows the approximation to capture multiple modes of the posterior. We demonstrate the efficacy of the nonparametric approximation with a hierarchical logistic regression model and a nonlinear matrix factorization model. We obtain predictive performance as good as or better than more specialized variational methods and MCMC approximations. The method is easy to apply to graphical models for which standard variational methods are difficult to derive.

## 1. Introduction

Approximate posterior inference—estimating the conditional distribution of hidden variables given some observations—is an important problem in many settings. In this paper, we develop a new variational inference algorithm for complex probabilistic mod-els. Compared to traditional variational methods, our method can capture more expressive distributions and be applied to a wider class of models.

Variational inference methods define some restricted family of distributions over the hidden variables $\theta$ and try to find the member of that family that is closest to the posterior. The family is chosen so that the problem of finding the distribution $q$ that best approximates the posterior becomes a tractable optimization problem.

Variational methods are effective and widely used. These methods usually find a unimodal approximation of the posterior, especially when the variational family is the commonly chosen mean-field family (Jordan et al., 1999; Beal, 2003). Such an approximation is inadequate when the posterior is multimodal. Furthermore, variational inference algorithms are challenging to derive for models that lack conditional distributions in tractable exponential families (i.e., for models without conditional conjugacy).

We develop a variational inference method for continuous hidden variables that captures multimodality and can be applied to many non-conjugate models. The variational family is a mixture of Gaussians, where the variational parameters are the locations and variances of each mixture component. This family of distributions resembles classical kernel density estimators from nonparametric statistics (Silverman, 1986). To approximate the variational objective function, we use Taylor series approximations of the log joint distribution and a bound on the entropy. We call this method *nonparametric variational inference* (NPV). In contrast to traditional unimodal variational distributions, the multiple components of the mixture can capture different aspects of the posterior.

While mixture approximations have been studied in the variational inference literature (Bishop et al., 1998;

Jaakola & Jordan, 1998), we develop this idea into a more generally applicable framework. (We discuss other approaches to mixture approximations in Section 4.) NPV is "general" in the sense that it is not tailored to a specific model, only requiring that the first and second derivatives of the log joint probability $\log p(\theta, y)$ be computable. Thus, it can be used in non-conjugate settings, i.e., where conditionals of the the individual hidden variables cannot be computed, such as in Bayesian models with non-conjugate priors. While previous methods for variational inference in non-conjugate models rely on mathematics tailored to the problem at hand, NPV is easily adapted to many settings.

In the following sections, we describe the variational objective function using this family and a general-purpose algorithm to approximately optimize it. We illustrate its performance on two models. First, we show that it performs as well in Bayesian logistic regression as the method of Jaakkola & Jordan (2000), which is tailored to that specific model. Second, we show that it outperforms several MCMC methods for a non-conjugate matrix factorization model of brain activity data (Gershman et al., 2011). Nonparametric variational inference is a promising strategy for approximating posterior distributions in complex probabilistic models.

## 2. Variational inference

We consider the problem of computing the posterior distribution of hidden variables $\theta \in \mathbb{R}^D$ given observed data $y$,

$$p(\theta|y) = \frac{p(y|\theta)p(\theta)}{p(y)}. \qquad (1)$$

This computation is analytically intractable for many models of interest because the denominator is difficult to compute.

The idea behind variational methods is to approximate $p(\theta|y)$ with a distribution $q(\theta)$ that belongs to a constrained family of distributions, indexed by a variational parameter (Jordan et al., 1999; Beal, 2003). The goal is to choose a member of that family that is "closest" to the posterior. In variational inference, closeness is measured by Kullback-Leibler (KL) divergence,

$$\mathrm{KL}[q(\theta)||p(\theta|y)] = \mathbb{E}_q\left[\log\frac{q(\theta)}{p(\theta|y)}\right]. \qquad (2)$$

Thus, inference becomes an optimization problem: we choose the variational parameter to minimize the KL divergence. The family of distributions is chosen to make this optimization tractable.

The KL divergence is difficult to optimize because it requires knowing the distribution that we are trying to approximate. In variational inference, we maximize an objective that is equal to the negative KL divergence plus a constant. Recall that $\mathrm{KL}[q(\theta)||p(\theta|y)] \geq 0$. We define a lower bound on the log marginal likelihood (evidence) $\log p(y)$ through the relation

$$\log p(y) = \mathcal{F}[q] + \mathrm{KL}[q(\theta)||p(\theta|y)], \qquad (3)$$

where

$$\mathcal{F}[q] = \mathbb{E}_q\left[\log\frac{p(y,\theta)}{q(\theta)}\right] = \mathcal{H}[q] + \mathbb{E}_q\left[f(\theta)\right] \qquad (4)$$

is the negative free energy, also known as the *evidence lower bound* (ELBO). Here $\mathcal{H}[q]$ is the entropy of $q$ and $f(\theta) = \log p(y, \theta)$. The ELBO is equal to the negative KL divergence plus the marginal distribution of the observations, which is constant with respect to the family $q$. It therefore reaches a maximum when $p(\theta|y) = q(\theta)$, where the KL is zero. Note that this is only attainable when the target posterior $p(\theta|y)$ is in the variational family, which it usually is not. Typically, $q$ will be constrained to a family of simpler distributions, and $\mathcal{F}[q]$ is optimized to find the distribution in this family that is closest (in KL) to the true posterior.

The most commonly used variational inference algorithm is mean-field variational inference. Mean-field methods find $q$ from the family of factorized posteriors: $q(\theta) = \prod_i q_i(\theta_i)$, where it is often convenient to choose $q_i(\theta_i)$ to have the same functional form as the conditional distribution $p(\theta_i|\theta_{-i}, y)$. When $p(\theta_i)$ is chosen to be conjugate to $p(y|\theta)$, the calculus of variations leads to closed-form coordinate ascent updates that converge to a local maximum of $\mathcal{F}[q]$ (Beal, 2003).

Despite the computational convenience of the mean-field approximation, it can be overly restrictive if there are strong dependencies between the hidden variables in the posterior distribution. Moreover, the closed-form updates are only available when using conjugate priors; many likelihood models of interest, such as logistic regression and the multilayer perceptron, cannot be paired with conjugate priors, making the application of mean-field methods more difficult.

## 3. Nonparametric variational inference

We now consider a flexible family of variational approximations that admits an efficient inference algorithm. Our algorithm is appropriate for models with continuous-valued hidden random variables, and does not require conjugacy between pairs of variables.

We choose the distribution $q(\theta)$ to be a uniformly-weighted Gaussian mixture with isotropic covariances,

$$q(\theta) = \frac{1}{N} \sum_{n=1}^{N} \mathcal{N}(\theta; \mu_n, \sigma_n^2 \mathbf{I}), \quad (5)$$

where $\mu_n$ is the mean of the $n$th Gaussian component and $\sigma_n^2$ is its variance. We call this a "nonparametric" family: We are making a weak set of assumptions about the shape of the posterior, since the Gaussian mixture family can approximate arbitrarily complex posteriors given a sufficient number of components. Further, this family resembles kernel density estimators used in classical nonparametric statistics (Silverman, 1986), with $\mu_n$ playing the role of a kernel center and $\sigma_n^2$ playing the role of a bandwidth parameter.

### 3.1. The Evidence Lower Bound

If $q$ is in the family defined by Eq. 5, we cannot compute the ELBO $\mathcal{F}[q]$; in general there is no closed-form expression either for the expectation of a nonlinear function under a Gaussian distribution or for the entropy of a mixture of Gaussians. However, we can approximate the ELBO and optimize this approximation (see Lawrence, 2000; Honkela et al., 2007, for other approaches to this problem). First, we lower bound the entropy term $\mathcal{H}[q]$. Then, we approximate the expected log joint $\mathbb{E}_q[\log p(y, \theta)]$.

We lower bound the entropy (the first term in Eq. 4) using Jensen's inequality (Huber et al., 2008),

$$\mathcal{H}[q] = -\int_\theta q(\theta) \log q(\theta) d\theta$$

$$= -\int_\theta q(\theta) \log \frac{1}{N} \sum_{n=1}^{N} \mathcal{N}(\theta; \mu_n, \sigma_n^2 \mathbf{I}) d\theta$$

$$\geq -\frac{1}{N} \sum_{n=1}^{N} \log \int_\theta q(\theta) \mathcal{N}(\theta; \mu_n, \sigma_n^2 \mathbf{I}) d\theta. \quad (6)$$

Each integral in Eq. 6 is the sum of $N$ convolved Gaussians, each component convolved with the $n$th. We obtain the final bound by using the fact that the convolution of two Gaussians is another Gaussian,

$$\mathcal{H}[q] \geq -\frac{1}{N} \sum_{n=1}^{N} \log q_n, \quad (7)$$

where $q_n = \frac{1}{N} \sum_{j=1}^{N} \mathcal{N}(\mu_n; \mu_j, (\sigma_n^2 + \sigma_j^2)\mathbf{I})$.

We now turn to the expected log joint $f(\theta)$, which is the second term in Eq. 4,

$$\mathbb{E}_q[f(\theta)] = \frac{1}{N} \sum_{n=1}^{N} \int_\theta \mathcal{N}(\theta; \mu_n, \sigma_n^2 \mathbf{I}) f(\theta) d\theta.$$

We approximate each term in this sum with a second-order Taylor series expansion of $f(\theta)$ around $\mu_n$,

$$f(\theta) \approx \hat{f}_n(\theta) = f(\mu_n) + \nabla f(\mu_n)(\theta - \mu_n) + \frac{1}{2}(\theta - \mu_n)^\top \mathbf{H}_n(\theta - \mu_n), \quad (8)$$

where $\mathbf{H}_n = \nabla_\theta^2 f(\mu_n)$ is the Hessian matrix of second derivatives. The approximate expectation is

$$\mathbb{E}_q[f(\theta)] \approx \frac{1}{N} \sum_{n=1}^{N} \int_\theta \mathcal{N}(\theta; \mu_n, \sigma_n^2 \mathbf{I}) \hat{f}_n(\theta) d\theta$$

$$= \frac{1}{N} \sum_{n=1}^{N} f(\mu_n) + \frac{\sigma_n^2}{2} \text{Tr}(\mathbf{H}_n). \quad (9)$$

This approximation is known as the multivariate delta method for moments (Bickel & Doksum, 2007), and is often used within variational inference schemes for models that cannot exploit conjugacy (e.g., Braun & McAuliffe, 2010).

Finally, we add the bound in Eq. 7 to the approximation in Eq. 9. This gives the approximate ELBO[1]

$$\mathcal{L}_2[q] = \frac{1}{N} \sum_{n=1}^{N} f(\mu_n) + \frac{\sigma_n^2}{2} \text{Tr}(\mathbf{H}_n) - \log q_n. \quad (10)$$

Intuitively, the likelihood term, $f(\mu_n)$, encourages placing samples in areas of high probability density, while the entropy term, $\log q_n$, penalizes "overcrowded" locations (i.e., where many samples are near each other). The Hessian term captures the local curvature of the posterior, discouraging the algorithm from placing samples in areas with high probability density but low volume (and therefore low mass).

We note two attractive properties of the approximate ELBO in Eq. 10. First, we have made no conjugacy assumptions; our only requirement is that the log joint $f(\theta) = \log p(\theta, y)$ is twice differentiable (or thrice differentiable if one wishes to use gradient ascent; but see below). Second, although the objective function involves a Hessian term, it only requires the calculation of the diagonal components; the cost of computing the diagonal of the Hessian is comparable to the cost of computing the gradient.

### 3.2. Optimizing the ELBO

Eq. 10 is a tractable approximation of the ELBO in Eq. 4. Our goal is now to maximize Eq. 10 with respect

---

[1] When some parameters are bounded, one can use nonlinear transformations to map an unbounded parameterization to a bounded range (e.g., the logistic function for variables in $[0, 1]$). In this case, one should add $\log |\mathbf{J}|$ to the approximate ELBO, where $\mathbf{J}$ is the Jacobian matrix of first derivatives of the transformation.

---

**Algorithm 1** Nonparametric variational inference

**Input:** data $y$, number of components $N$.
**Initialize** $\theta_{1:N}$ randomly.
**repeat**
    **for** $n = 1$ **to** $N$ **do**
        $\mu_n \leftarrow \text{argmax}_{\mu_n} \mathcal{L}_1[q]$.
    **end for**
    $\sigma_{1:N}^2 \leftarrow \text{argmax}_{\sigma_{1:N}^2} \mathcal{L}_2[q]$.
**until** change of $\mathcal{L}_2[q]$ is less than 0.0001.

---

to the variational parameters $\mu_n$ and $\sigma_n$. One option is to use a gradient-based solver. However, there is a serious computational problem with this approach— computing the gradient of Eq. 10 requires computing a matrix of *third* derivatives, since we must compute the gradient of the Hessian trace $\text{Tr}(\mathbf{H}_n)$. This leads to a cost that is quadratic in the number of parameters.

To avoid the calculation of third derivatives, we use both first- and second-order approximations of the ELBO. The first-order approximation is

$$\mathcal{L}_1[q] = \frac{1}{N}\sum_{n=1}^{N} f(\mu_n) - \log q_n. \quad (11)$$

This is obtained in the same way as Eq. 10, but using a first-order approximation of $f(\theta)$, rather than the second-order approximation in Eq. 8. We iterate between optimizing the variances $\sigma$ using the second-order approximation in Eq. 10 and optimizing the means $\mu$ using the Eq. 11. Each optimization is done using L-BFGS. We found that it is more efficient to optimize $\mathcal{L}_1[q]$ with respect to one mean at a time, holding the others fixed, and iterating over components. This coordinate ascent procedure converges faster than batch optimization of $\mu_{1:N}$, but coordinate and batch optimization produce similar results. Our algorithm is summarized in Algorithm 1.

Both $\mathcal{L}_1[q]$ and $\mathcal{L}_2[q]$ are approximations of $\mathcal{F}[q]$. Splitting the optimization problem into these two steps allows us to avoid the cost of calculating the gradient of $\frac{\sigma_n^2}{2}\text{Tr}(\mathbf{H}_n)$ with respect to the means $\mu$. In our experiments, 3 iterations typically proved sufficient to achieve convergence. Although the first-order approximation may appear drastic, it still achieves our main goal: placing kernels in areas of high probability mass. Further simulation work is needed to assess the trade-offs involved in this approximation.

As an illustration, we constructed a synthetic multimodal "posterior" $f(\theta)$ using a mixture of skewed bivariate $t$-distributions. Figure 1 shows $f(\theta)$ alongside the NPV approximation with several settings of $N$.

With $N = 1$, the approximation is only able to capture a single mode, but with $N = 2$ it is able to capture the two modes with high fidelity, though it cannot capture the true covariance structure or the heavy tails. With $N = 10$, the approximation better captures the skew by placing several low-variance components along the diagonal. This illustration demonstrates some strengths and weaknesses of the NPV approximation: it can capture multi-modality, but the isotropic covariance of the components makes it difficult to capture skew in the posterior. This problem can be ameliorated by using more components.

Note that the number of parameters that need to be fit with NPV increases linearly with $N$ (the number of components in the mixture). This may pose challenges for models with a large number of hidden variables. On the other hand, it may only be necessary to use a small number of components (e.g., less than 10) to capture the major aspects of the posterior (as suggested by Figure 1). We note also that the KL divergence between the mixture distribution $q$ and the true posterior decreases at best logarithmically in the number of mixture components $N$, suggesting that there may be diminishing returns to using very large values of $N$ (Jaakola & Jordan, 1998).

### 3.3. Relationship to other algorithms

The NPV objective relates to several other methods. When there is one component $N = 1$, the entropy term $\log q_1$ does not depend on the mean $\mu_1$, and when $\sigma_1^2$ becomes sufficiently small, the Hessian term of Eq. 10 goes to 0. Consequently, the NPV objective when $N = 1$ and $\sigma_1 \to 0$ is

$$\mathcal{L}[q] = \log p(y, \mu) + \text{const.} = \log p(\theta = \mu | y) + \text{const.}$$

The maximum of this function is the maximum a posteriori (MAP) solution.

When $N = 1$ and $\sigma_1^2$ is allowed to vary, we obtain a Gaussian approximation centered around the MAP solution. This can be understood as a diagonalized *Laplace approximation* (MacKay, 1995), i.e., where we ignore correlations between the dimensions of $\theta$. The Laplace approximation has drawbacks: for example, it is not invariant to reparameterization, it performs badly when the mean and mode of the posterior are far apart, and it cannot capture multiple modes (Beal, 2003).

When $N > 1$ and $\sigma_n^2 \to 0$, we obtain a quasi-Monte Carlo approximation of the posterior, $q(\theta) = \frac{1}{N}\sum_{n=1}^{N} \delta_{\mu_n}(\theta)$, where $\delta_{\mu_n}(\cdot)$ is a Dirac point mass located at $\mu_n$. Thus one way to look at the NPV algorithm is as a deterministic sampling method.

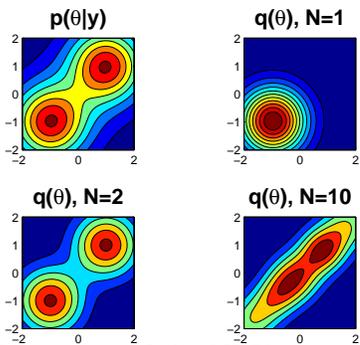

*Figure 1.* **Illustration of the NPV approximation fit to a multimodal posterior**. The true posterior (*top left*), constructed as a mixture of bivariate *t*-distributions, alongside the NPV approximation fit with several settings of $N$.

## 4. Related work

Approximate inference for non-conjugate models is an active area of research. Some authors have used numerical or Monte Carlo methods to approximate intractable integrals. For example, Lawrence et al. (2004) used importance sampling to approximate the expectations required for inference in a Bayesian model of microarray images. Ihler et al. (2009) generalized particle filtering for approximate inference in factor graphs with continuous variables. Honkela et al. (2007) used numerical quadrature to approximate expectations in a nonlinear factor analysis model. These techniques are useful, but may fail in high dimensions.

Several researchers use specialized approximations for certain classes of models, such as those with logistic nonlinearities (e.g., Jaakkola & Jordan, 2000; Khan et al., 2010). In contrast, our goal is to develop an algorithm for inference in general non-conjugate models with continuous hidden variables.

Closely related to our method is the *mixture mean-field* (MMF) method (Bishop et al., 1998; Jaakkola & Jordan, 1998; Lawrence, 2000), which models the posterior as a mixture of mean-field approximations. Recently, Bouchard & Zoeter (2009) revisited this approach using soft-binning functions. NPV can be viewed as a special case of MMF because each component factorizes into a collection of one-dimensional Gaussian sub-components (due to the isotropic covariances). Our innovation is that we exploited the functional form of the Gaussian mixture to derive an efficient approximate inference algorithm. NPV requires no user input beyond specifying the joint likelihood function, its gradient, optionally the diagonal of its Hessian, and the number of components. These modest requirements give NPV a practical advantage in situations where it is difficult to derive the MMF up-dates.

## 5. Applications

In this section, we apply the NPV algorithm to several probabilistic models and compare its performance to other widely-used methods.

### 5.1. Logistic regression

In this section, we ask whether NPV produces reasonable approximations for models where closed-form updates can be applied. We focus on a hierarchical logistic regression model and compare its accuracy to a standard variational treatment (Jaakkola & Jordan, 2000, henceforth "JJ").

**Generative model**. The observed data $y = \{\mathbf{c}, \mathbf{X}\}$ consist of $T$ binary class labels, $c_t \in \{-1, 1\}$, and $K$ covariates for each datapoint, $\mathbf{x}_t \in \mathbb{R}^K$. The hidden variables $\theta = \{\mathbf{w}, \alpha\}$ consist of $K$ regression coefficients $w_k \in \mathbb{R}$, and a precision parameter $\alpha \in \mathbb{R}_+$. We assume the following model (MacKay, 1995):

$$p(\alpha) = \text{Gamma}(\alpha; a, b) \tag{12}$$

$$p(w_k|\alpha) = \mathcal{N}(w_k; 0, \alpha^{-1}) \tag{13}$$

$$p(c_t = 1|\mathbf{x}_t, \mathbf{w}) = \frac{1}{1 + \exp(-\mathbf{w}^\top \mathbf{x}_t)}. \tag{14}$$

Here $a$ and $b$ are hyperparameters (shape and inverse scale, respectively) that we assume to be fixed.

**Results**. We evaluated NPV and JJ on 13 binary classification data sets compiled by Mika et al. (1999).[2] The number of covariates in these data sets ranges from 2 to 60, and the number of observations ranges from 24 to 7400. We used split-half training/testing. We used the following hyperparameter settings: $a = 1$, $b = 0.01$, $N = 5$ (similar results were obtained with $N = 10$).

The predictive distribution for NPV was approximated using a Monte Carlo estimate. We drew 1000 samples from the fitted variational mixture of Gaussians and estimated the log-likelihood of the test data as an average of the log-likelihoods under each sample. Figure 2 (top) compares the log-likelihood of the test data under the NPV and JJ approximations. NPV and JJ achieve statistically indistinguishable accuracy. Figure 2 (bottom) shows the same comparison for the ELBO, confirming that NPV closely mimics the JJ approximation. We emphasize that JJ exploits special properties of the generative model (i.e., a clever lower bound on the logistic sigmoid function), whereas NPV only uses

---

[2] http://theoval.cmp.uea.ac.uk/matlab/default.html

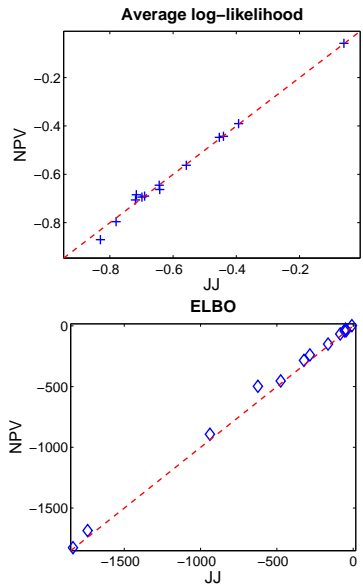

**Average log-likelihood**

**ELBO**

*Figure 2.* **Accuracy of NPV for logistic regression**. (*Top*) Each axis represents the log-likelihood of predictions for test data under NPV or the Jaakkola & Jordan (2000) algorithm (JJ), conditional on the inputs **X**. Each point represents one of 13 data sets compiled by Mika et al. (1999). For NPV, $N = 5$ components were used (similar results were obtained with $N = 10$). (*Bottom*) Same as above for the ELBO.

the derivatives of the joint distribution.

We also fit the model using an MCMC algorithm, Hamiltonian (or Hybrid) Monte Carlo (HMC; Neal, 2011), which takes the same inputs as NPV (the log joint probability and its gradient). HMC uses the gradient of $f(\theta)$ to efficiently explore the posterior, making it one of the most effective samplers for models with continuous variables. With 1000 samples, we found that this algorithm predicts held-out data significantly worse ($p < 0.00001$, Wilcoxon signed-rank test) compared to NPV and JJ. Presumably the inferior performance of HMC could be improved by running the sampler for longer, but this would result in greater computational overhead.

### 5.2. Topographic latent source analysis

We now study our method with a more complicated model, for which standard variational algorithms are inapplicable. We apply the NPV approximation to a nonlinear latent variable model of functional magnetic resonance imaging (fMRI) data. Data from fMRI experiments contain measurements of brain activity that are collected while a subject performs a task, such as labeling images. The goal of these experiments is to understand the relationship between cognitive pro-cesses and brain activity. One reason this problem is complicated is that fMRI data is spatial. Brain activity is measured in 3D brain-space (a grid of "voxels"). Measurements made on nearby voxels are dependent.

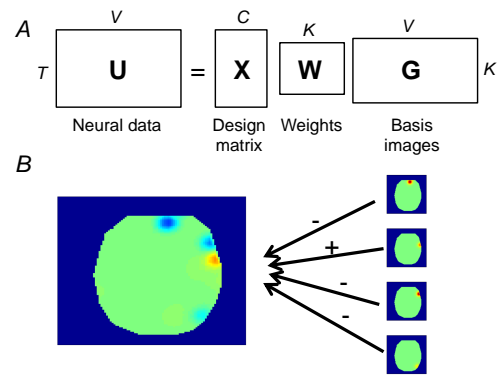

*Figure 3.* **Schematic of the topographic latent source analysis model.** (*A*) Matrix factorization view of the data-generating process; (*B*) illustration of how latent sources combine to produce neural data.

Gershman et al. (2011) developed a factorization model of spatial patterns in fMRI data, *topographic latent source analysis* (TLSA). TLSA decomposes voxel activations into a set of spatial functions (topographic latent sources). These functions are related to task and cognitive variables (called "covariates") through a weight matrix that is also inferred from the data. We can evaluate the quality of a fitted model by using it to predict held-out brain data, conditional on covariates. Unlike traditional probabilistic matrix factorization models, TLSA is not conditionally conjugate and closed-form mean-field inference is not available. Gershman et al. (2011) approximated the posterior with MCMC, but their method was too slow to analyze large data sets.

**Generative model.** Each datapoint $t$ in an fMRI experiment consists of a vector of $V$ voxel activations, $\mathbf{u}_t \in \mathbb{R}^V$, and a vector of $C$ covariates, $\mathbf{x}_t \in \mathbb{R}^C$. The intuition behind TLSA is that the spatial organization of voxel activations arises from a small number of anatomically localized brain regions involved in processing the task. Formally, TLSA decomposes the voxel activations into a covariate-dependent superposition of $K$ latent sources:

$$u_{tv} = \sum_{c=1}^{C} x_{tc} \sum_{k=1}^{K} w_{ck} g_{kv} + \epsilon_{tv}, \qquad (15)$$

where $\epsilon_{tv} \sim \mathcal{N}(0, \tau^{-1})$ is a Gaussian noise term, $w_{ck}$ is a weight that specifies how covariate $c$ influences source $k$, and $g_{kv}$ is the activation of source $k$ in voxel $v$. This generative process (illustrated in Figure 3) can be viewed as a probabilistic matrix factorization model where $\{\mathbf{g}_k\}$ are basis images that are combined to produce the observed neural activity.

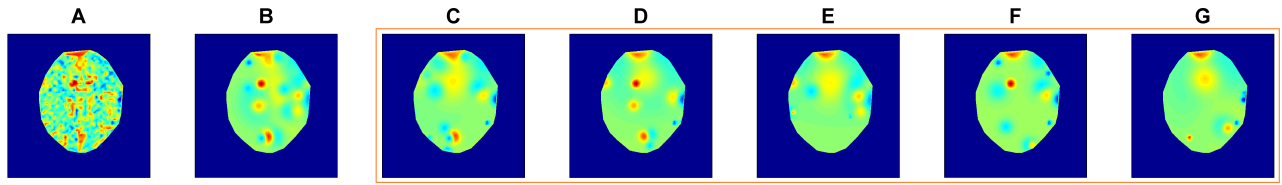

*Figure 4.* **Neural data and reconstructions**. (*A*) Average test image. (*B*) MAP reconstruction. (*C-G*) Reconstructions for each component of the NPV approximation. Notice that each component captures idiosyncrasies (corresponding to different local optima of the objective function) missed by the MAP reconstruction.

Each basis image is constructed by evaluating a parameterized spatial basis function at each voxel location. Following Gershman et al. (2011), we chose this function to be a radial basis function with parameters $\omega_k = \{\bar{\mathbf{r}}_k, \lambda_k\}$:

$$g_{kv} = \exp\left\{-\lambda_k^{-1}||\mathbf{r}_v - \bar{\mathbf{r}}_k||^2\right\}, \tag{16}$$

where $\bar{\mathbf{r}}_k \in [0,1]^M$ is the source center (in normalized coordinates), $\lambda_k \in \mathbb{R}_+$ is a width parameter, and $\mathbf{r}_v \in [0,1]$ is the location of voxel $v$. In the notation of Section 2, the observed variables are $y = \{\mathbf{X}, \mathbf{U}, \mathbf{R}\}$ and the hidden variables are $\theta = \{\mathbf{W}, \mathbf{G}\}$.

To complete the generative model, we placed the following priors on the parameters:

$$w_{ck} \sim \mathcal{N}(0, \sigma_w^2), \quad \bar{r}_{kd} \sim \text{Beta}(1,1), \quad \lambda_k \sim \text{Exp}(\rho).$$

In all our analyses, we used the following hyperparameter settings: $\tau = 1, \sigma_w^2 = 5, \rho = 1$.

**Results**. We fit TLSA to data collected by Mason and Just (unpublished), involving subjects viewing words. Each word was either the name of a type of tool or of a type of building (i.e., there were 2 classes), and the subject's task was to think about the word and its properties. There were a total of 84 trials per subject (see Gershman et al., 2011, for more details). We restricted our analysis to a 1,323 voxels (a single slice of the brain activity data) from a single subject.

We trained the model on one half and then generated predictions of the neural data for the other half, conditioning on the test covariates. For NPV, we approximated the predictive distribution using a Monte Carlo estimate, as described in the previous section. As an illustration of the model fits, Figure 4 shows the average data and reconstructions derived from the MAP estimate and the NPV approximation. While the MAP estimate captures the global pattern of activity, each component of the NPV approximation captures small idiosyncrasies that may be difficult to extract using a single point estimate. In other words, the NPV approximation captures several local maxima of the posterior; we next show that this translates into better predictive accuracy.

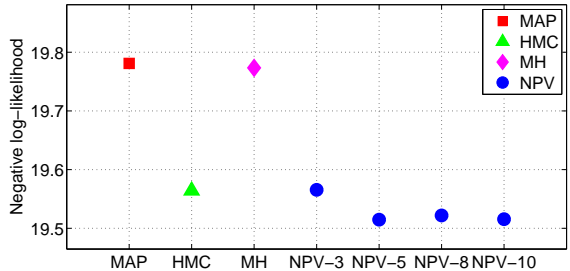

*Figure 5.* **The nonparametric variational approximation improves TLSA predictions of held-out data**. The Y-axis represents the negative log-likelihood of predictions for held-out neural data, conditional on the covariates. In all cases $K = 20$ sources were used. Standard error bars are smaller than the markers.

We evaluated the quality of the reconstruction by calculating the mean-squared reconstruction error of held-out neural data, a quantity proportional to the negative log-likelihood of the held-out data. We also fit TLSA using HMC (see above); we collected 5000 samples, keeping the last 200 for the predictive distribution. We repeated this procedure for the Metropolis-Hastings (MH) sampler used in the original TLSA paper (Gershman et al., 2011). The results are shown in Figure 5. NPV works well with a varying number of components (though best when $N > 3$), substantially outperforming the MAP and MCMC estimators.

We re-emphasize here that TLSA is non-conjugate, and hence MMF cannot be applied without using specially-tailored approximations (Lawrence, 2000). Note that while both MH and HMC are asymptotically guaranteed to perfectly approximate the posterior, these algorithms require tuning and are often slow to converge. In our experiments, NPV was about 3 times faster than HMC.

## 6. Discussion

We developed an approximate inference method for posteriors that do not necessarily enjoy the conjugacy properties that make common variational approximations (e.g., mean-field) possible. Our algorithm is easy to apply to new probabilistic models; all that is required is the likelihood function and its gradient (a requirement shared by many other algorithms, including MAP estimation and HMC). When applied to a hierarchical logistic regression model, we found that NPV incurs little loss in accuracy compared to a more specialized variational algorithm (Jaakkola & Jordan, 2000). We further showed, using a nonlinear latent variable model of fMRI data, that NPV can find an approximation of the posterior that improves predictive performance over MAP estimation and MCMC.

NPV has limitations. First, it assumes a simple approximating family. This could be improved by introducing a full covariance matrix into the component distributions or by allowing the components to be non-uniformly weighted. Further, NPV only applies to continuous variables. We plan to extend it to models with discrete hidden variables.

In summary, NPV is a posterior inference algorithm that is a step towards generically applicable variational approximations. The need for such approximations is increasing, as researchers begin to explore more and more complicated probabilistic models to cope with the increasing complexity of large data sets. Our hope is that by employing generic inference algorithms, the hard work of inference can proceed "invisibly," and researchers can devote more time to testing and refining the assumptions of their models.

**Acknowledgements.** Samuel J. Gershman is supported by a graduate research fellowship from the NSF. Matthew D. Hoffman is supported by NSF ATM-0934516, DOE DE-SC0002099, and IES R305D100017. David M. Blei is supported by ONR N00014-11-1-0651, NSF CAREER 0745520, NSF IIS-1009542, AFOSR FA9550-09-1-0668, the Alfred P. Sloan foundation, and a grant from Google.